\documentclass{article}

\usepackage{arxiv}

\usepackage[utf8]{inputenc} 
\usepackage[T1]{fontenc}    
\usepackage{hyperref}       
\usepackage{url}            
\usepackage{booktabs}       
\usepackage{amsfonts}       
\usepackage{nicefrac}       
\usepackage{microtype}      
\usepackage{lipsum}
\usepackage{graphicx}

\usepackage{times}
\usepackage{latexsym}

\usepackage[T1]{fontenc}

\usepackage[utf8]{inputenc}

\usepackage{microtype}

\author{
	Amirmohammad Omidi Galdiani\thanks{These authors contributed equally.}\\
	Department of Computer Engineering\\
	University of Guilan\\
	Rasht, Guilan, Iran\\
	\texttt{amirmohammadomidi@proton.me} \\
	\And
	Sepehr Rezaei Melal\footnotemark[1]\\
	Department of Computer Engineering\\
	University of Guilan\\
	Rasht, Guilan, Iran\\
	\texttt{sepehrrezaeimelal@gmail.com} \\
	\And
	Mohammad Norasteh\footnotemark[1]\\
	Department of Computer Engineering\\
	University of Guilan\\
	Rasht, Guilan, Iran\\
	\texttt{houmannoraste80@gmail.com} \\
	\And
	Arash Yousefi Jordehi\\
	Department of Computer Engineering\\
	University of Guilan\\
	Rasht, Guilan, Iran\\
	\texttt{arashy76@phd.guilan.ac.ir} \\
	\And
	Seyed Abolghasem Mirroshandel\\
	Department of Computer Engineering\\
	University of Guilan\\
	Rasht, Guilan, Iran\\
	\texttt{mirroshandel@guilan.ac.ir} \\
}

\usepackage{minted}
\usepackage{mdframed}

\surroundwithmdframed[
backgroundcolor=white,  
linecolor=white,        
]{minted}

\setminted[text]{
	fontsize=\small,
	breaklines,
	style=default,         
}

\usepackage{inconsolata}
\usepackage{xcolor}

\usepackage{booktabs}
\usepackage{soul}

\usepackage[acronym,nomain]{glossaries}
\newacronym{nlp}{NLP}{Natural Language Processing}
\newacronym{om}{OM}{Opinion Mining}
\newacronym{crf}{CRF}{Conditional Random Fields}

\newacronym{dl}{DL}{Deep Learning}
\newacronym{fgom}{FGOM}{Fine-Grained Opinion Mining}
\newacronym{pe}{PE}{Prompt Engineering}
\newacronym{bert}{BERT}{Bidirectional Encoder Representations from Transformers}
\newacronym{t5}{T5}{Text-to-Text Transfer Transformer}
\newacronym{mpqa}{MPQA}{Multi-Perspective Question Answering}
\newacronym{ml}{ML}{Machine Learning}
\newacronym{ese}{ESE}{Expressive Subjective Element}
\newacronym{ds}{DS}{Direct Subjective}
\newacronym{ose}{OSE}{Objective Speech Event}
\newacronym{nltk}{NLTK}{Natural Language Toolkit}
\newacronym{lstm}{LSTM}{Long Short Term Memory}
\newacronym{bilstm}{BiLSTM}{Bi-directional Long Short Term Memory}
\newacronym{cnn}{CNN}{Convolutional Neural Network}
\newacronym{rnn}{RNN}{Recurrent Neural Network}
\newacronym{llrd}{LLRD}{Layer-wise Learning Rate Decay}
\newacronym{mtl}{MTL}{Multi-Task Learning}
\newacronym{json}{JSON}{JavaScript Object Notation}
\newacronym{html}{HTML}{Hyper Text Markup Language}
\newacronym{xml}{XML}{eXtensible Markup Language}
\newacronym{sota}{SOTA}{state-of-the-art}
\newacronym{LM}{LM}{Language Model}
\newacronym{LLM}{LLM}{Large Language Model}
\newacronym{cl}{CL}{Curriculum Learning}
\newacronym{PL}{PL}{Progressive Learning}
\newacronym{DA}{DA}{Data Augmentation}
\newacronym{CV}{CV}{Computer Vision}
\newacronym{AI}{AI}{Artificial Intelligence}
\newacronym{al}{AL}{Active Learning}
\newacronym{SCL}{SCL}{Supervised Contrastive Learning}
\newacronym{GPT}{GPT}{Generative Pre-trained Transformer}
\newacronym{cv}{CV}{Computer Vision}
\newacronym{knn}{KNN}{K-Nearest Neighbors}
\newacronym{STS}{STS}{Semantic Textual Similarity}
\newacronym{crossv}{CV}{Cross-Validation}
\newacronym{srl}{SRL}{Semantic Role Labeling}
\newacronym{orl}{ORL}{Opinion Role Labeling}
\newacronym{apc}{APC}{automatic prompt construction}

\usepackage[colorinlistoftodos,textsize=tiny]{todonotes}

\usepackage{arydshln}
\usepackage{csquotes} 
\usepackage{amssymb}

\newcommand{\numInstances}{97,169 }

\title{From Semantic Roles to Opinion Roles: SRL Data Extraction for Multi-Task and Transfer Learning in Low-Resource ORL}

\begin{document}
	\maketitle
	\begin{abstract}
		This report presents a detailed methodology for constructing a high-quality Semantic Role Labeling (SRL) dataset from the Wall Street Journal (WSJ) portion of the OntoNotes 5.0 corpus and adapting it for Opinion Role Labeling (ORL) tasks. Leveraging the PropBank annotation framework, we implement a reproducible extraction pipeline that aligns predicate-argument structures with surface text, converts syntactic tree pointers to coherent spans, and applies rigorous cleaning to ensure semantic fidelity. The resulting dataset comprises \numInstances predicate-argument instances with clearly defined Agent (ARG0), Predicate (REL), and Patient (ARG1) roles, mapped to ORL's Holder, Expression, and Target schema. We provide a detailed account of our extraction algorithms, discontinuous argument handling, annotation corrections, and statistical analysis of the resulting dataset. This work offers a reusable resource for researchers aiming to leverage SRL for enhancing ORL, especially in low-resource opinion mining scenarios.
	\end{abstract}
	\section{Introduction}
	
	\gls*{orl} aims to identify the \textbf{Holder}, \textbf{Expression}, and \textbf{Target} of subjective statements in text. Despite its importance in fine-grained opinion mining, \gls*{orl} remains a low-resource task, with only a few annotated datasets, such as MPQA, available for training robust models. In contrast, \gls*{srl} benefits from large, richly annotated corpora like OntoNotes 5.0, which provide detailed predicate-argument structures grounded in linguistic theory. The close structural alignment between \gls*{srl} and \gls*{orl}, where \gls*{srl}'s \textbf{Agent (ARG0)}, \textbf{Predicate (REL)}, and \textbf{Patient (ARG1)} correspond closely to \gls*{orl}'s roles, creates a strong opportunity for effective \textbf{knowledge transfer} from \gls*{srl} to \gls*{orl}.
	
	Although several studies have explored \gls*{srl}-to-\gls*{orl} transfer, most have described their pipelines only at a high level, omitting critical implementation details. This lack of transparency makes it difficult for other researchers to replicate, validate, or adapt their methods. The aim of this report is to address this gap by providing a \textbf{comprehensive, step-by-step methodology} for extracting \gls*{srl} data from the Wall Street Journal portion of OntoNotes 5.0, mapping it to the \gls*{orl} schema, and preparing it as a high-quality dataset ready for downstream use. We describe each stage in detail, from syntactic pointer resolution and discontinuous argument handling to role mapping, cleaning, and export, ensuring that every step can be reproduced without ambiguity.
	
	The resulting dataset is designed for \textbf{broad applicability} in modern \gls*{nlp} workflows. In \textbf{transfer learning}, it can serve as a semantic foundation for models trained in low-resource \gls*{orl} settings. In \textbf{multi-task learning}, it can be integrated alongside \gls*{orl} data to promote shared representation learning between semantic and opinion roles. In \textbf{active learning}, it can provide a large, high-quality pool from which the most relevant \gls*{srl} instances can be selected for annotation in \gls*{orl} tasks. By releasing both the dataset and the full extraction code in a public GitHub repository, we aim to provide a reusable, extensible resource for the research community.
	
	This technical report is therefore intended not only as a presentation of results, but also as a \textbf{clear, replicable blueprint} for bridging \gls*{srl} and \gls*{orl} in resource-constrained opinion mining scenarios.
	The complete extraction code, preprocessing scripts, and final dataset are publicly released at \url{https://github.com/Amirmohammadomidi/srl-data-extraction}.
	
	\section{Contributions}
	
	This technical report makes the following key contributions:
	\begin{enumerate}
		\item \textbf{A fully documented \gls*{srl} extraction pipeline}. We present a reproducible, detailed method for parsing PropBank annotations, resolving syntactic pointers, handling discontinuous arguments, and mapping \gls*{srl} roles to the \gls*{orl} schema.
		\item \textbf{A high-quality \gls*{srl} dataset aligned to \gls*{orl} roles}. We produce a dataset of \numInstances predicate-argument instances from OntoNotes 5.0 (WSJ), with ARG0, REL, and ARG1 mapped to Holder, Expression, and Target respectively.
		\item \textbf{Error correction and cleaning procedures}. We identify and correct known OntoNotes annotation errors, remove syntactic traces, and normalize text for compatibility with modern \gls*{nlp} models.
		\item \textbf{Support for multiple learning paradigms}. The dataset is designed to be directly usable in \textbf{transfer learning}, \textbf{multi-task learning}, and \textbf{active learning} settings for low-resource \gls*{orl}.
		\item \textbf{Open-source code and resources}. We release all extraction scripts, dataset preparation tools, and documentation in a public GitHub repository, enabling full reproducibility and community extension.
	\end{enumerate}

	\section{Related Research} \label{sec:related_research} 
	
	Research in \gls*{om} can be broadly categorized into four interconnected directions, each addressing distinct aspects of opinion analysis. 
	
	The \textbf{first direction} focuses on \textit{opinion expression extraction and labeling}, which aims to identify textual spans that explicitly convey opinions. Traditional methods relied on tagging schemes such as BMESO combined with \gls*{crf} models to capture fine-grained opinion spans~\cite{breck2007identifying, yang-cardie-2013-joint, yang2014joint}. To better model contextual dependencies, researchers integrated Recurrent Neural Networks, notably BiLSTM-CRF architectures~\cite{katiyar-cardie-2016-investigating}. More recently, contextualized embeddings from models like BERT and ELMo have further improved performance by enriching token representations with deep semantic context~\cite{quan2019end}. 
	
	The \textbf{second direction} addresses \textit{end-to-end opinion structure recognition}, where systems simultaneously detect opinion expressions and their associated roles (Holder and Target). This unified approach mitigates error propagation inherent in pipeline methods. Early neural architectures, such as transition-based parsers, effectively captured both expressions and roles~\cite{ZHANG201956, wu2022mastering}. Later developments leveraged BERT’s contextual embeddings integrated with BiLSTM-CRF layers~\cite{quan2019end}, while dependency graph convolutional networks incorporated syntactic information into opinion labeling processes, resulting in significant performance enhancements~\cite{zhang-etal-2020-syntax}. 
	
	The \textbf{third direction} concerns \textit{\gls*{orl} in a given-expression setting}, where the opinion expression is pre-identified, and the task reduces to labeling its semantic roles: Holder and Target. Given the limited size of labeled \gls*{orl} datasets, \gls*{mtl} strategies have emerged as effective solutions.  \textnormal{Marasović} and Frank~\cite{marasovic-frank-2018-srl4orl} demonstrated that \gls*{srl}, a well-resourced task with strong structural parallels to \gls*{orl}, can serve as an effective auxiliary signal in \gls*{mtl} frameworks, substantially boosting \gls*{orl} performance in low-resource conditions. 
	
	The \textbf{fourth direction} encompasses \textit{comprehensive, unified frameworks for opinion mining} that integrate multiple subtasks—such as expression detection, role labeling, and relation classification—into a single cohesive model. \cite{xia-etal-2021-unified} introduced a span-based \gls*{mtl} system leveraging syntactic constituents to address both end-to-end and given-expression \gls*{orl} tasks within one framework. While effective, this approach faced high computational complexity due to the need to process all possible spans. Wu et al.~\cite{wu2022mastering} addressed this by proposing a neural transition-based model combined with PointNets, enhancing boundary detection while relying heavily on external syntactic knowledge. Most recently, Yousefi Jordehi et al.~\cite{yousefi2024improving} proposed the Generative Opinion Mining (GenOM) system, a T5-based generative framework that unifies both settings through Multi-Task Learning and Automatic Prompt Construction (APC). Their APC mechanism automatically optimizes prompts during training, eliminating the need for manual prompt engineering and achieving state-of-the-art results on MPQA without relying on external syntactic or semantic knowledge. This work represents a significant shift toward fully generative, prompt-aware architectures for holistic opinion analysis. 
	
	Despite these advances, critical challenges remain. Traditional sequence-labeling approaches are often insufficient for handling nested and overlapping opinion structures. Generative models, while addressing structural complexity, require effective strategies for leveraging external semantic and syntactic knowledge. Furthermore, data scarcity continues to impede \gls*{orl} model generalizability, motivating ongoing research in multi-task and transfer learning methodologies. 
	
	In summary, research on \gls*{om} has progressively evolved from basic tagging methods toward sophisticated, unified frameworks that better reflect the nuanced structure of subjective language. These developments underscore the importance of modeling complex relational structures and reinforce the need for transparent, reproducible methodologies to support further innovation in low-resource settings. 

	\section{SRL Dataset Construction from OntoNotes 5.0}
	
	This section presents a fully reproducible pipeline for extracting a high-quality \gls*{srl} dataset from the Wall Street Journal (WSJ) portion of OntoNotes 5.0. The resulting dataset comprises \numInstances predicate-argument instances with ARG0, REL, and ARG1 roles, explicitly mapped to \gls*{orl}'s Holder, Expression, and Target schema.
	
	\subsection{Data Sources and Pre-requisites}
	
	The pipeline integrates two core resources:
	
	\begin{itemize}
		\item \textbf{PropBank Release:} We utilize the PropBank release (available at \url{https://github.com/propbank/propbank-release.git}), which provides semantic role annotations layered on Penn Treebank syntax. We focus on the \texttt{data/ontonotes/nw/wsj} directory containing .prop files.
		\item \textbf{OntoNotes 5.0 Corpus:} The OntoNotes 5.0 corpus supplies the underlying syntactic trees and plain text sentences. Files with \texttt{.onf} and \texttt{.parse} extensions are accessed through shared resources via direct download links.
		
	\end{itemize}
	
	The \texttt{semantic\_role\_labeling.py} script initiates the process by cloning the PropBank repository and navigating to the WSJ annotation directory.
	
	\subsection{Extraction of PropBank Propositions}
	
	The core of PropBank annotation resides in \texttt{.prop} files, each encoding predicate-argument structures for individual sentences. The extraction process systematically identifies and parses these files:
	
	\begin{itemize}
		\item \textbf{File Identification:} The script iterates over folder numbers (\texttt{folder\_num}), ranging from `00` to `24`, and zero-padded file numbers (\texttt{file\_num}). File paths are constructed using f-strings:
		\begin{verbatim}
			f"{path}/{prefix}_{sub_file}.prop"
		\end{verbatim}
		For example: \path{propbank-release/data/ontonotes/nw/wsj/00/wsj_0001.prop}.
		Figure~\ref{fig:wsj_0001_prop_content} displays the content of such a file.
		
		\begin{figure*}[htb]
			\centering
			\includegraphics[width=0.99\linewidth]{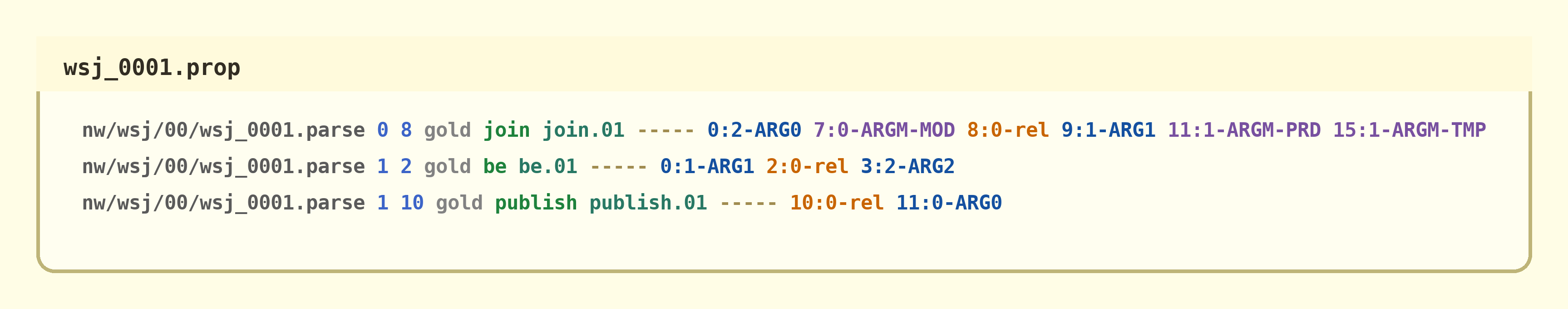}
			\caption{A sample \texttt{.prop} file and its contents: \texttt{wsj\_0001.prop}.}
			\label{fig:wsj_0001_prop_content}
		\end{figure*}
		
		\item \textbf{File Parsing:} \texttt{os.path.exists()} verifies file presence. Existing files are opened in read mode, and content is read into \texttt{data} before appending to the \texttt{sample} list.
		
		\item \textbf{Line Splitting:} Each \texttt{sample} entry is split on newline characters (\texttt{\textbackslash n}), and lines are collected into \texttt{lst}.
		
		\item \textbf{Proposition Extraction:} 
		Each \texttt{.prop} file contains multiple predicate–argument structures separated by newline characters (\texttt{\textbackslash n}). 
		During processing, the script reads the content of each file and splits it based on newline separators, resulting in individual proposition strings. 
		Each list of propositions is then appended as a sub-list to the main \texttt{split\_annotations} list. 
		As a result, \texttt{split\_annotations} becomes a list of lists, where each sub-list represents the propositions contained in a single WSJ file. 
		This process yields 1,724 sub-lists in total, corresponding to the range of files from \texttt{wsj\_0001.prop} to \texttt{wsj\_2454.prop}, as illustrated in Figure~\ref{fig:structure_split_annotations_list}.
		
		\begin{figure*}[htb]
			\centering
			\includegraphics[width=0.99\linewidth]{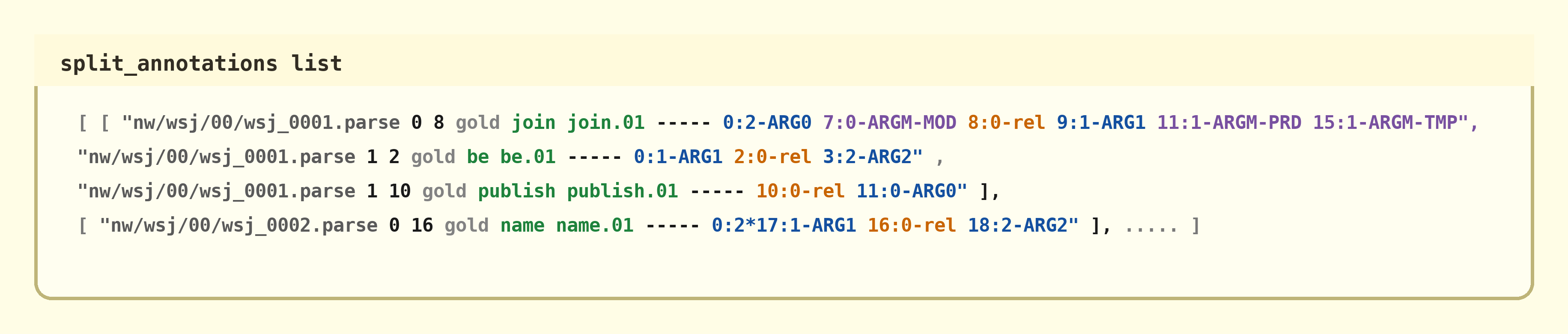}
			\caption{Hierarchical structure of \texttt{split\_annotations}: each sub-list contains all raw PropBank propositions from a single WSJ file.}
			\label{fig:structure_split_annotations_list}
		\end{figure*}
		
		\item \textbf{Proposition Sorting:} The helper function \texttt{num\_sort} extracts integers via \texttt{re.findall(r'\textbackslash d+', s)} and returns indices 2--4 (exclusive) as a sorting key for \texttt{split\_annotations[i].sort(key=num\_sort)}.
		
		\item \textbf{Data Cleaning (File Removal):} Four entries—at indices 112, 612, 617, and 621—are removed from \texttt{split\_annotations} and \texttt{existing\_files} due to structural discrepancies with the GitHub repository.
		
		\item \textbf{Predicate-Tree Index Mapping:} The \texttt{index} list maps each predicate to its syntactic tree index. For each proposition in \texttt{split\_annotations[i]}, the second token (\texttt{.split()[1]}) is extracted and converted to an integer. These indices are accumulated into the global \texttt{index} list, as shown in Figure~\ref{fig:structure_of_the_index_list}.
		\begin{figure}[htb]
			\centering
			\includegraphics[width=0.36\linewidth]{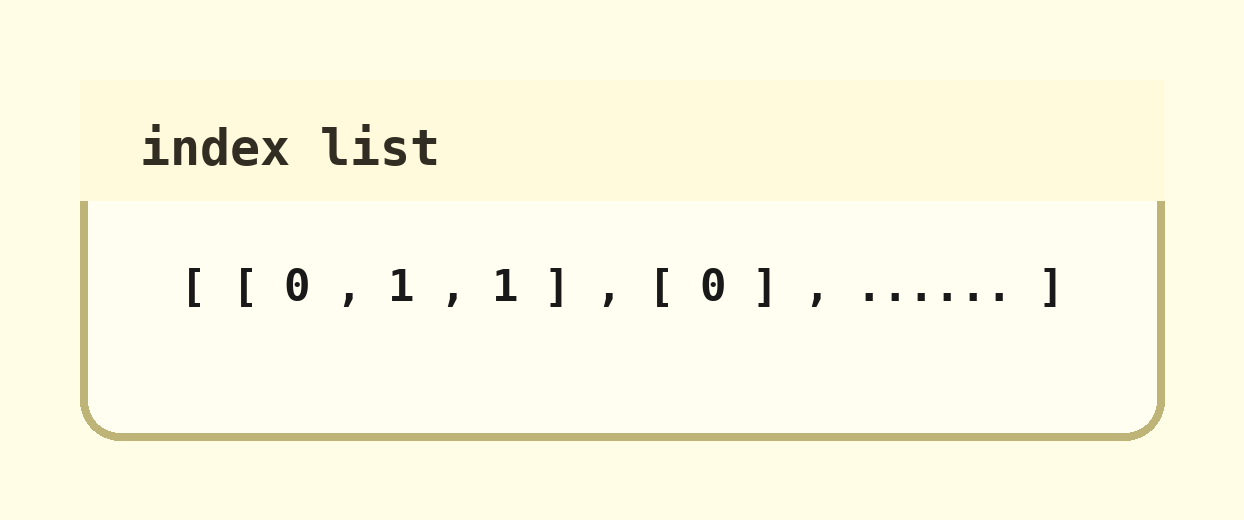}
			\caption{Structure of \texttt{index} list: mapping each predicate (proposition) to its corresponding parse tree index within the file.}
			\label{fig:structure_of_the_index_list}
		\end{figure}
	\end{itemize}
	
	\subsection{Extraction of Sentences and Trees}
	
	To align propositions with surface text and syntactic structure, sentences and parse trees are extracted from \texttt{.onf} files using the previously identified \texttt{existing\_files}:
	
	\begin{itemize}
		\item \textbf{Sentence Extraction (Standard and with Anchors):} File content is split on double newlines. Sentence blocks are identified by long hyphen sequences (section delimiters), as shown in Figure~\ref{fig:sample_onf_file_its_contents_compact_form_wsj_0001_onf}.
		\begin{figure*}[htb]
			\centering
			\includegraphics[width=0.8\linewidth]{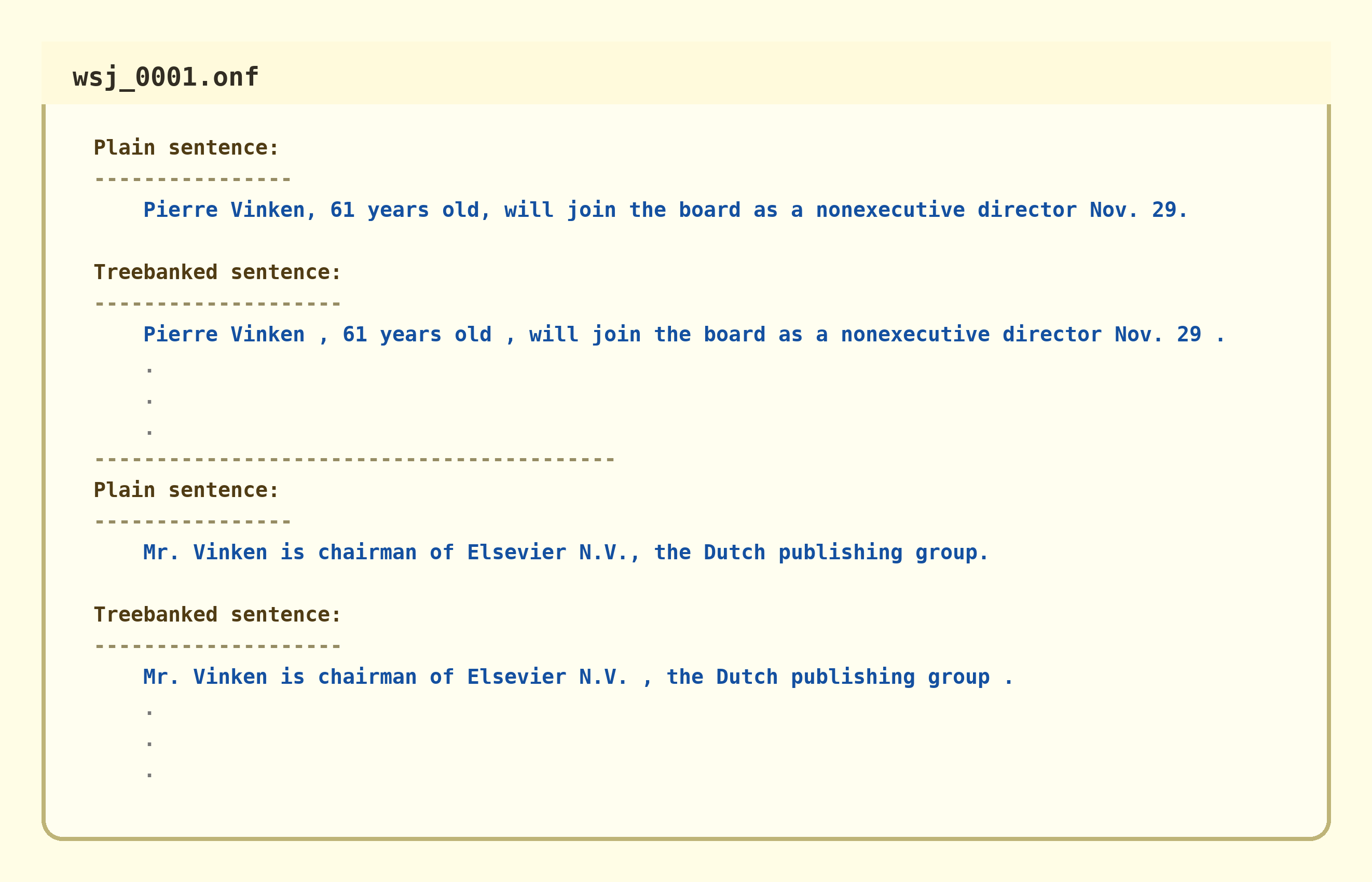}
			\caption{Compact view of \texttt{wsj\_0001.onf}: the standard sentence appears after the hyphen delimiter, followed by the anchored version containing trace markers (e.g., \texttt{*PRO*-1}).}
			\label{fig:sample_onf_file_its_contents_compact_form_wsj_0001_onf}
		\end{figure*}
		Two variants are extracted:
		\begin{itemize}
			\item \texttt{sent}: Standard text (without trace markers).
			\item \texttt{sent\_anchor}: Text with trace anchors (e.g., \texttt{*PRO*-1}).
		\end{itemize}
		
		The script reads each \texttt{.onf} file, splits its content by double newline characters (\texttt{\textbackslash n\textbackslash n}), 
		and locates sentence blocks using long sequences of hyphens as delimiters. 
		From each file, two lists are extracted:
		\begin{itemize}
			\item \texttt{split\_standard\_sentences}: Standard (plain) sentences without traces.
			\item \texttt{split\_sentences\_with\_anchor}: Treebanked sentences containing trace anchors such as \texttt{*PRO*-1}.
		\end{itemize}
		
		The hierarchical structure of \texttt{split\_standard\_sentences} is depicted in Figure~\ref{fig:structure_content_of_split_standard_sentences_list}.
		\begin{figure*}[htb]
			\centering
			\includegraphics[width=0.8\linewidth]{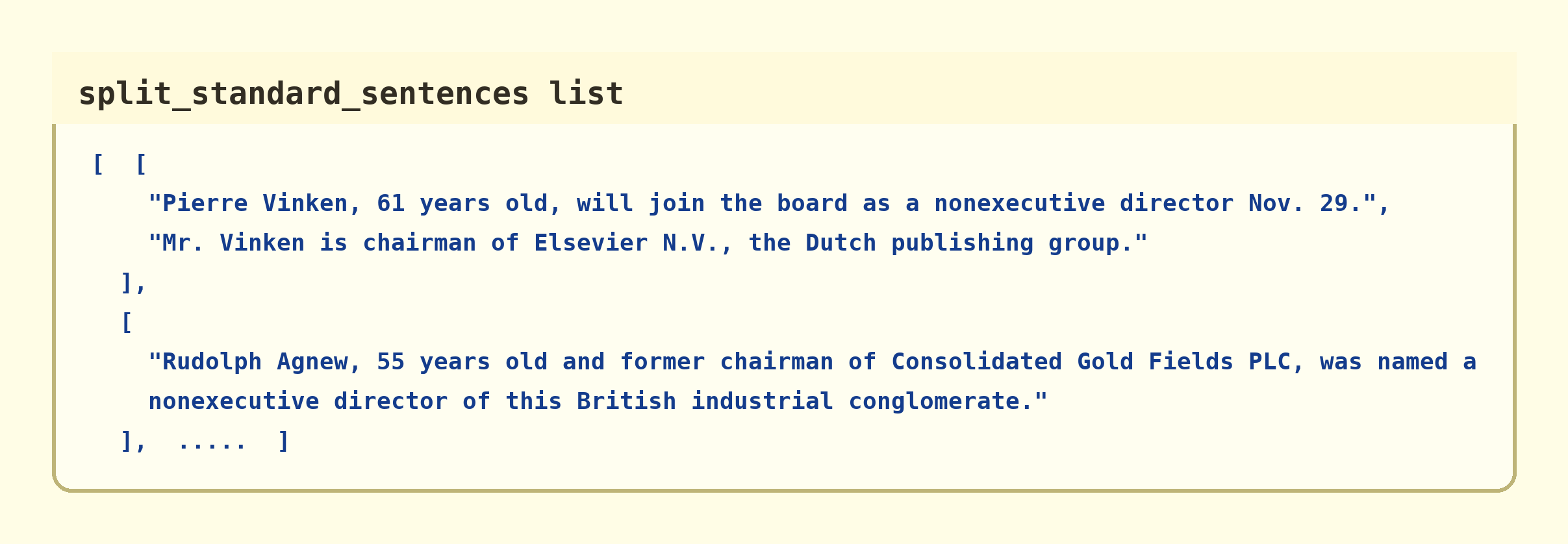}
			\caption{Hierarchical organization of \texttt{split\_standard\_sentences}: each sub-list contains all standard sentences from one \texttt{.onf} file.}
			\label{fig:structure_content_of_split_standard_sentences_list}
		\end{figure*}
		
		\item \textbf{Tree Extraction:} 
		The script extracts syntactic parse trees from files with the \texttt{.parse} extension, following the same segmentation logic used for sentence extraction. 
		Each \texttt{.parse} file contains one or more trees corresponding to the sentences within its paired \texttt{.onf} file. 
		During processing, each file is read and split based on double newline characters (\texttt{\textbackslash n\textbackslash n}), which separate individual parse trees. 
		The resulting tree strings are then appended as sub-lists to the main \texttt{split\_trees} list, ensuring that each sub-list represents the collection of parse trees from a single WSJ file. 
		An example of this structure is illustrated in Figure~\ref{fig:structure_and_content_of_parse_files_a_sample_of_wsj_0001_parse}.
		
		\begin{figure}[htb]
			\centering
			\includegraphics[width=0.51\linewidth]{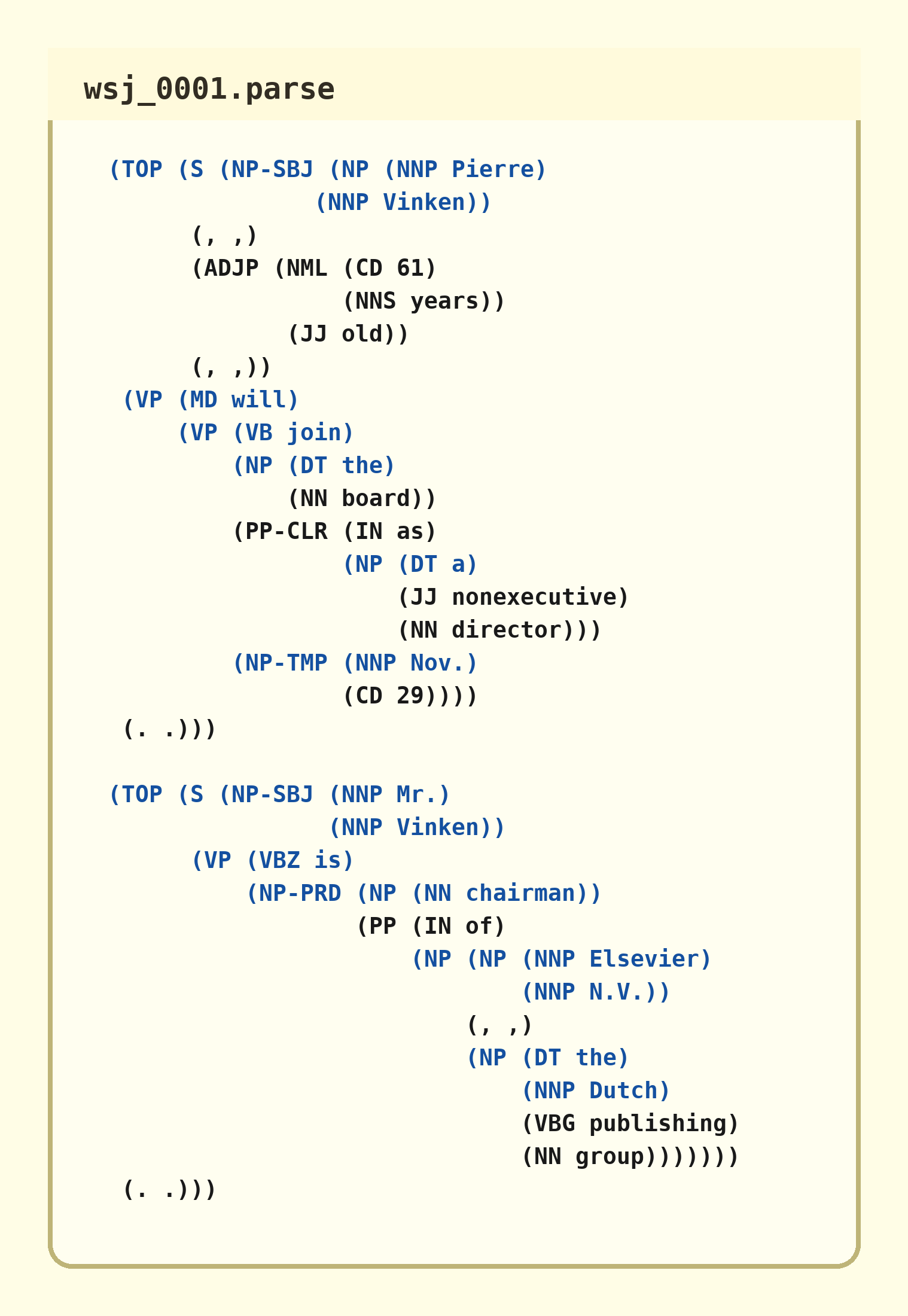}
			\caption{Structure of \texttt{wsj\_0001.parse}: two parse trees separated by double newlines, ready for alignment with propositions.}
			\label{fig:structure_and_content_of_parse_files_a_sample_of_wsj_0001_parse}
		\end{figure}
	\end{itemize}
	
	\subsection{Annotation Extraction and Conversion}
	
	The next phase extracts role annotations (ARG0, ARG1, REL) from raw proposition strings and converts them to textual spans:
	
	\begin{itemize}
		\item \textbf{Annotation Extraction:} Three lists are initialized: \texttt{arg\_zero} (ARG0), \texttt{arg\_one} (ARG1), and \texttt{rel}. For each proposition in the flattened \texttt{split\_annotations}:
		\begin{itemize}
			\item The string is split into tokens.
			\item Tokens ending in \texttt{-ARG0}, \texttt{-ARG1}, or \texttt{-rel} are identified, and the prefix (e.g., \texttt{0:1} from \texttt{0:1-ARG0}) is extracted.
		\end{itemize}

		The content and structure of \texttt{arg\_zero}, \texttt{arg\_one}, and \texttt{rel}	lists are shown in Figures~\ref{fig:9}, ~\ref{fig:10} and~\ref{fig:11}, respectively.
		\begin{figure}[htb]
			\centering
			\includegraphics[width=0.5\linewidth]{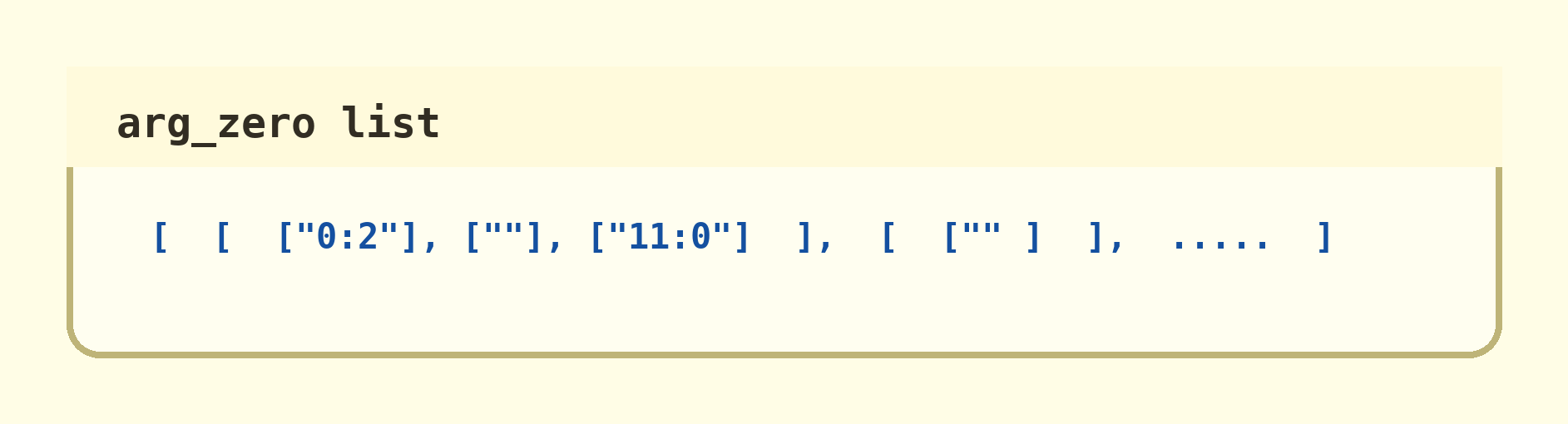}
			\caption{The content and structure of the \texttt{arg\_zero} list.}
			\label{fig:9}
		\end{figure}
		
		\begin{figure}[htb]
			\centering
			\includegraphics[width=0.55\linewidth]{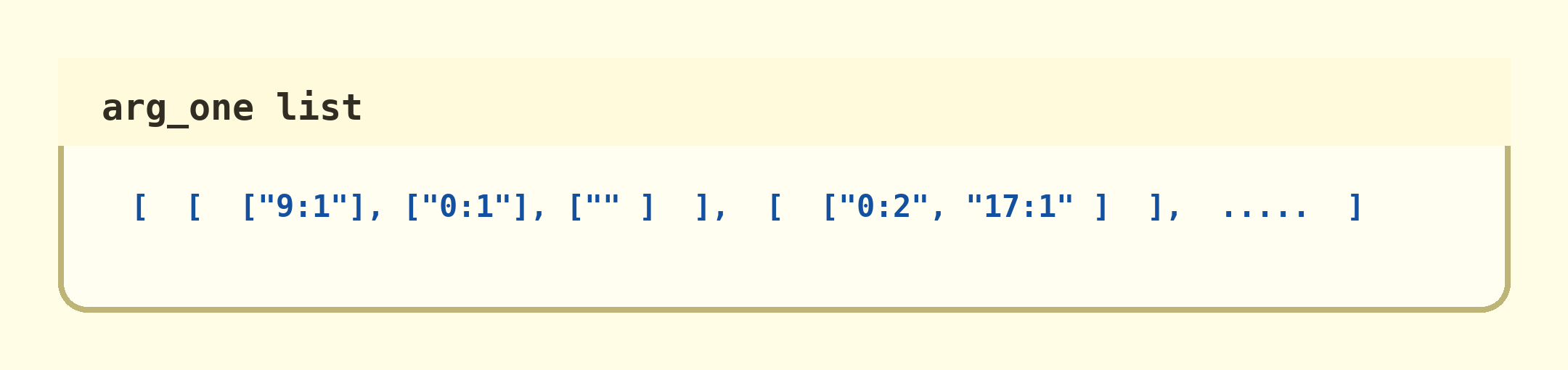}
			\caption{The content and structure of the \texttt{arg\_one} list.}
			\label{fig:10}
		\end{figure}
		
		\begin{figure}[htb]
			\centering
			\includegraphics[width=0.5\linewidth]{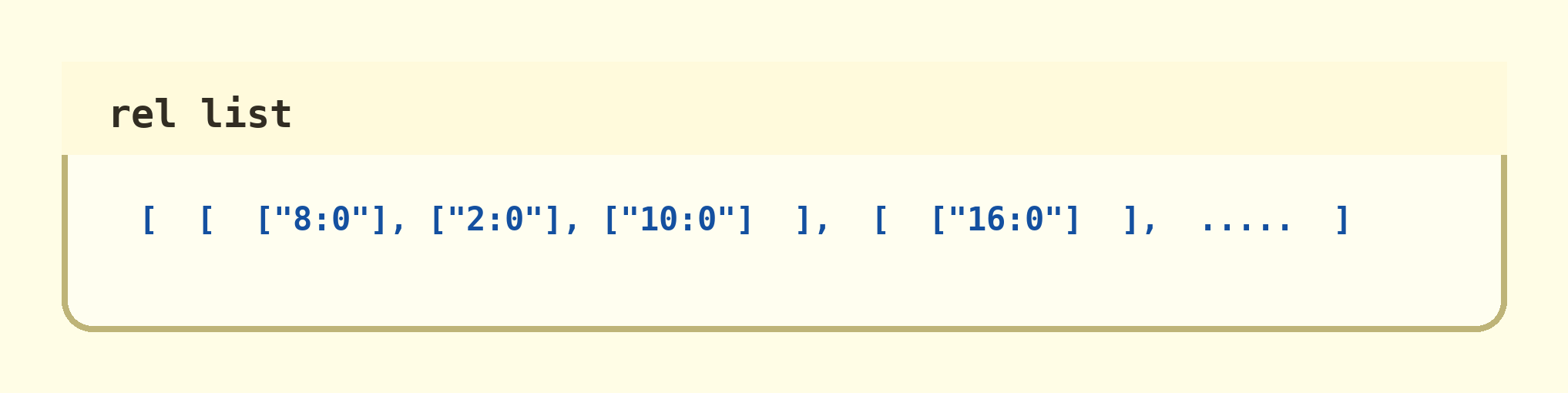}
			\caption{The content and structure of the \texttt{rel} list.}
			\label{fig:11}
		\end{figure}

		\item \textbf{Conversion to Tree Pointers:} The \texttt{TreePointer} function converts pointer strings (e.g., \texttt{0:2}) to \texttt{nltk.corpus.reader.propbank.PropbankTreePointer} objects using regex-based splitting on \texttt{:} or \texttt{;}.
		
		\item \textbf{Conversion to Text Spans:}
		This stage uses the \texttt{parse\_tree\_to\_text} function, which converts PropbankTreePointers into readable textual spans.
		It first converts the target subtree to an \texttt{nltk.tree.Tree} object using \texttt{Tree.fromstring()}, 
		applies the \texttt{.select()} method of the TreePointer, 
		and finally extracts the leaf tokens with \texttt{.leaves()} before joining them into surface text.

		\begin{itemize}
			\item Converts tree string to \texttt{nltk.tree.Tree} via \texttt{Tree.fromstring()}.
			\item Extracts leaf nodes using \texttt{.leaves()}.
			\item Joins tokens with spaces.
		\end{itemize}
		Its implementation is shown in Figure~\ref{fig:12}.
		\begin{figure*}[htb]
			\centering
			\includegraphics[width=0.65\linewidth]{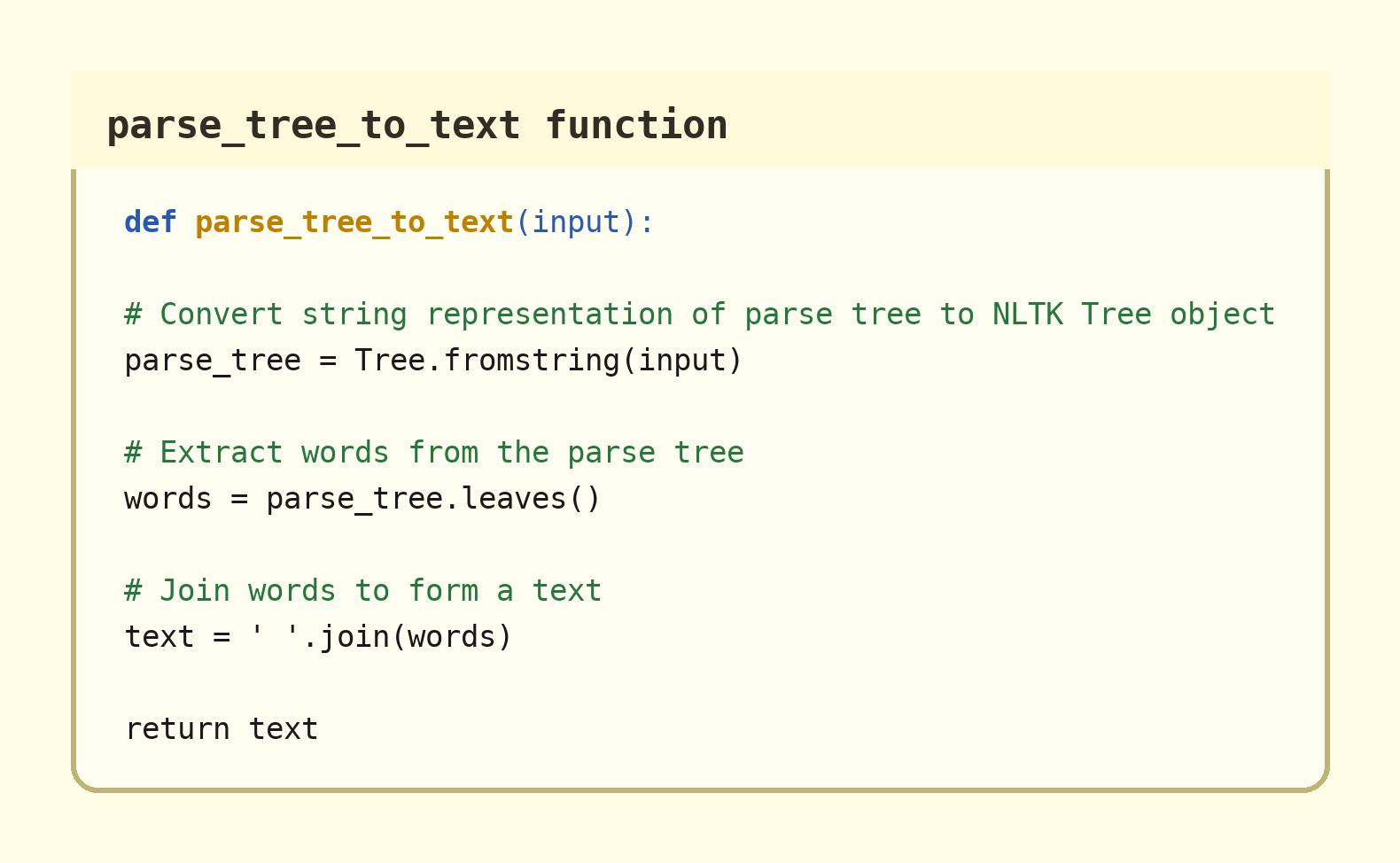}
			\caption{Core implementation of \texttt{parse\_tree\_to\_sentence}: converts a syntactic subtree into its surface text representation.}
			\label{fig:12}
		\end{figure*}
		For each non-empty annotation in \texttt{list\_ARG0s}, \texttt{list\_ARG1s}, and \texttt{list\_predicates}:
		\begin{itemize}
			\item \texttt{TreePointer} creates a pointer object.
			\item \texttt{.select()} extracts the subtree from \texttt{Tree[i]}.
			\item \texttt{parse\_tree\_to\_sentence} converts the subtree to surface text.
		\end{itemize}
		
		
		\item \textbf{Handling Chain Annotations:} The \texttt{converter} function processes discontinuous spans (e.g., \texttt{14:1*16:1*17:1-ARG0}) by splitting on \texttt{*}, \texttt{,}, or \texttt{;}, returning individual pointer strings.
	\end{itemize}
	
	\subsection{DataFrame Creation and Cleaning}
	
	The extracted data lists, including plain sentences, Treebanked sentences, predicates, and argument spans, are organized into a structured pandas \texttt{DataFrame} for efficient storage and analysis.
	
	\begin{itemize}
		\item \textbf{Trace Anchor Removal:} 
		Before creating the \texttt{DataFrame}, all trace anchors and special markers (for example, \texttt{*PRO*-2}, \texttt{*}, or numerical traces) are removed from the Treebanked sentences. 
		This cleaning step employs regular expressions to identify and eliminate all trace patterns, ensuring that the sentences reflect their true surface forms.
		
		\item \textbf{DataFrame Construction:} 
		A single \texttt{DataFrame} is constructed using the core extracted lists as columns:
		\begin{itemize}
			\item \texttt{list\_sentences}: Plain (standard) sentences. 
			\item \texttt{list\_sentences\_with\_anchor}: Treebanked sentences containing traces.
			\item \texttt{list\_predicates}: Predicate (verb) spans.
			\item \texttt{list\_ARG0s}: ARG0 text spans.
			\item \texttt{list\_ARG1s}: ARG1 text spans.
		\end{itemize}
		
		\item \textbf{Merged Argument Column:}
		A new feature column, \texttt{Merged\_Arguments}, is added to combine ARG0 and ARG1 spans within each row using a vertical bar separator (\texttt{ARG0|ARG1}). 
		This concatenation allows for compact representation and easier alignment with the \gls*{orl} schema, where both arguments are jointly used for Holder–Target mapping.
		
		\item \textbf{Filtering Incomplete Entries:}
		Rows with missing core arguments (where \texttt{Merged\_Arguments == "|"}) are removed from the dataset. 
		This ensures that only instances containing at least one of the core arguments (ARG0 or ARG1) remain. 
		After cleaning and index resetting, the resulting \texttt{DataFrame} contains \textbf{97,169} valid predicate–argument observations.
		
		\item \textbf{Final Structure:}
		The resulting structure includes five primary columns: plain sentence, Treebanked sentence, predicate span, ARG0 span, ARG1 span, and one derived column (\texttt{Merged\_Arguments}). 
		Figure~\ref{fig:13} illustrates the organized structure of the cleaned dataset.
		\begin{figure*}[!htb]
			\centering
			\includegraphics[width=\linewidth]{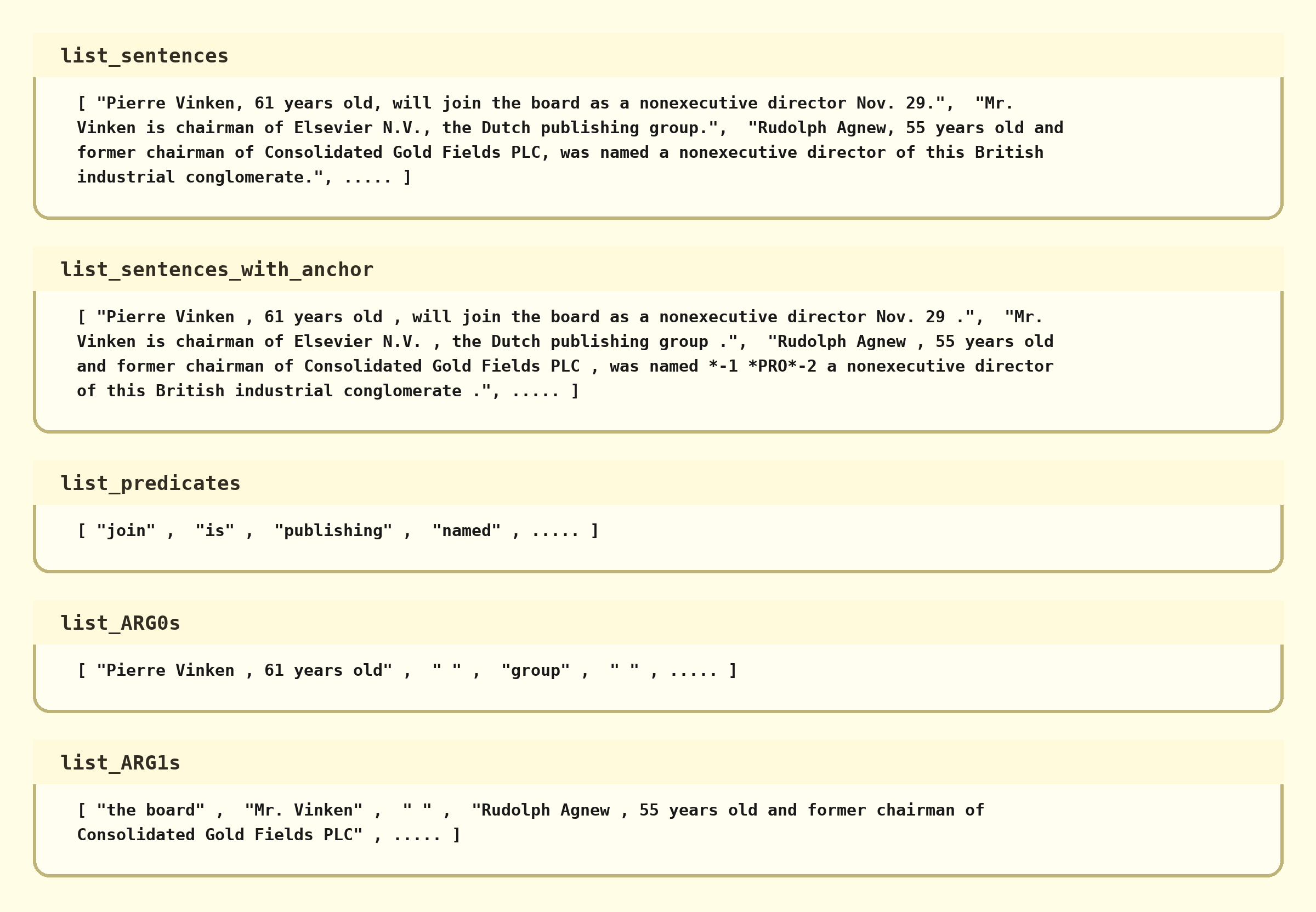}
			\caption{Final extracted and cleaned data components: plain sentences, predicates (REL), and argument spans for ARG0 and ARG1.}
			\label{fig:13}
		\end{figure*}
	\end{itemize}
	
	\subsection{Dataset Export}
	
	The finalized dataset, containing 97,169 cleaned predicate–argument instances, is exported to a CSV file named \texttt{dataset.csv}. 
	This export preserves full alignment between predicates and their corresponding ARG0–ARG1 spans, ensuring that each record can be directly used for downstream tasks such as transfer or multi-task learning.
	
	%
	
	\section{Dataset Statistics}
	
	The final dataset contains \numInstances high-quality predicate--argument instances extracted from the WSJ portion of OntoNotes 5.0. 
	The updated distribution of core arguments is as follows:
	
	\begin{itemize}
		\item \textbf{Both ARG0 and ARG1:} 52.1\%
		\item \textbf{Only ARG1:} 41.1\%
		\item \textbf{Only ARG0:} 6.8\%
	\end{itemize}
	
	\subsection{Argument Presence Breakdown}
	
	A detailed breakdown of argument co-occurrence across all instances is presented in Table~\ref{tab:arg_breakdown}. 
	More than half of the predicates (52.1\%) are annotated with both ARG0 and ARG1, forming complete semantic triples. 
	A further 41.1\% include only ARG1, often reflecting passive or eventive constructions without an explicit agent, 
	while 6.8\% contain only ARG0, typically corresponding to intransitive or stative clauses. 
	
	\begin{table}[htb]
		\centering
		\caption{Argument presence breakdown across \numInstances predicate instances.}
		\label{tab:arg_breakdown}
		\begin{tabular}{lc}
			\toprule
			Argument Configuration & Percentage (\%) \\
			\midrule
			Both ARG0 \& ARG1      & 52.1 \\
			Only ARG1              & 41.1 \\
			Only ARG0              & 6.8 \\
			\bottomrule
		\end{tabular}
	\end{table}
	
	\subsection{Predicate Distribution and Argument Length}
	
\begin{figure}[htb]
		\centering
		\includegraphics[width=0.7\linewidth]{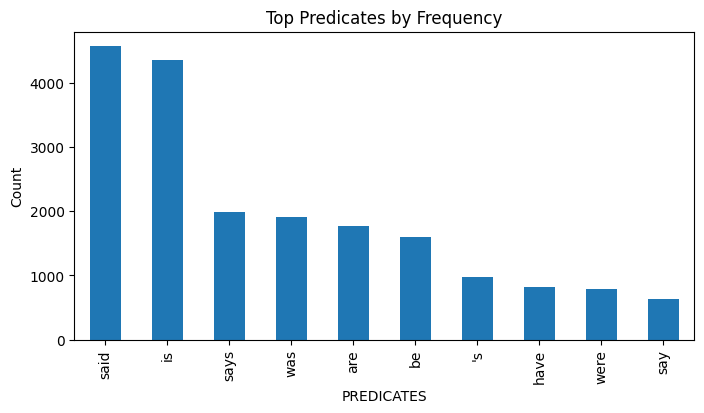}
		\caption{Top predicates by frequency in the dataset. The distribution exhibits a clear long-tail pattern dominated by reporting verbs.}
		\label{fig:top_predicates}
	\end{figure}

	The predicate vocabulary follows a long-tail distribution, with a small set of high-frequency verbs dominating usage. 
	As illustrated in Figure~\ref{fig:top_predicates}, the most frequent predicate is \texttt{said} (approximately 4,500 occurrences), 
	followed by \texttt{is} (around 4,300) and \texttt{says} (around 2,000). 
	Other frequent verbs include \texttt{was}, \texttt{are}, \texttt{be}, \texttt{have}, \texttt{were}, and \texttt{say}, 
	reflecting the typical reporting-oriented language of the Wall Street Journal corpus.
	
	\begin{figure}[htb]
		\centering
		\includegraphics[width=0.4\linewidth]{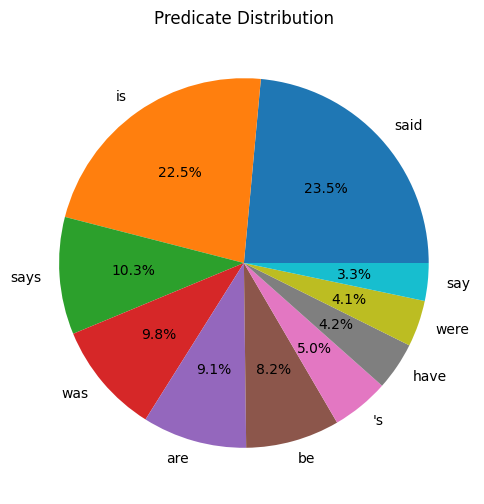}
		\caption{Proportional distribution of the ten most frequent predicates.}
		\label{fig:predicate_distribution}
	\end{figure}
	
	Analysis of argument span lengths reveals a structural asymmetry: ARG0 spans average 2.2 words, 
	typically corresponding to concise agent phrases, whereas ARG1 spans average 6.6 words, 
	reflecting longer and more complex patient or theme expressions. 
	This observation is consistent with PropBank annotation practices and supports robust modeling of target expressions 
	in low-resource \gls*{orl} settings.
	
	\begin{figure}[htb]
		\centering
		\includegraphics[width=0.45\linewidth]{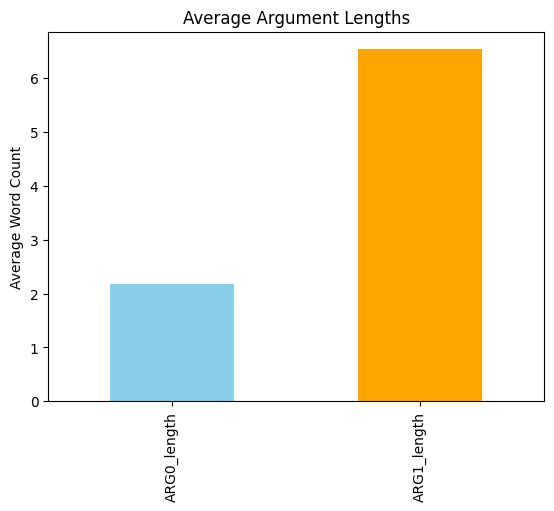}
		\caption{Average argument span lengths for ARG0 and ARG1.}
		\label{fig:arg_lengths}
	\end{figure}
	
	\subsection{Predicate Sentiment Analysis and Cross-domain Applicability}
	
	To investigate the semantic and affective tendencies of the extracted predicates, 
	we applied sentiment scoring using the VADER (Valence Aware Dictionary and sEntiment Reasoner) model. 
	Each predicate received a \texttt{compound} score in the range $[-1, 1]$, quantifying its polarity. 
	For categorical interpretation, scores were divided into five sentiment classes: 
	very negative ($-2$), negative ($-1$), neutral ($0$), positive ($1$), and very positive ($2$). 
	
\begin{figure}[htb]
		\centering
		\includegraphics[width=0.77\linewidth]{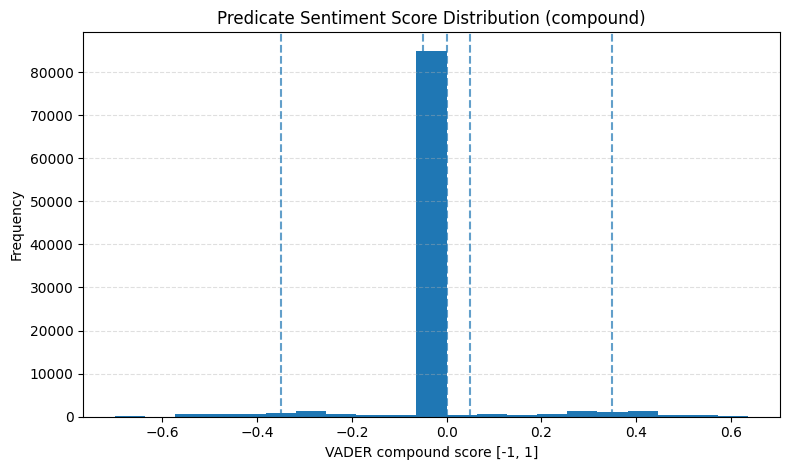}
		\caption{Distribution of predicate sentiment scores (\texttt{compound}) across all verbs. 
			Vertical dashed lines mark sentiment boundaries at $-t_2$, $-t_1$, $t_1$, and $t_2$.}
		\label{fig:predicate_sentiment_distribution}
	\end{figure}

\begin{figure}[htb]
		\centering
		\includegraphics[width=0.73\linewidth]{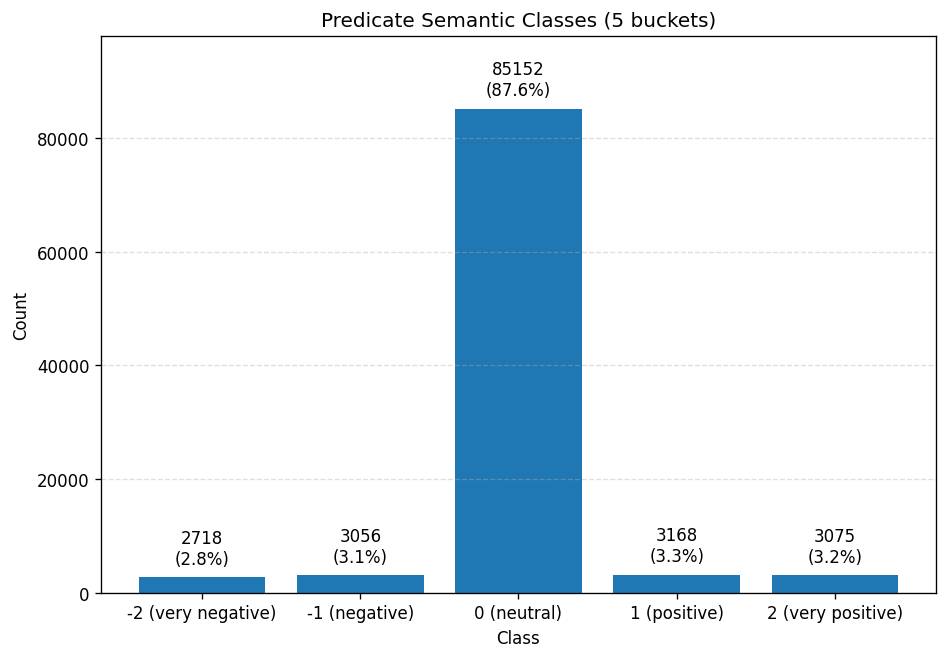}
		\caption{Frequency of predicates across five sentiment classes. 
			The majority of verbs exhibit neutral polarity, with minor portions showing positive or negative tone.}
		\label{fig:predicate_sentiment_buckets}
	\end{figure}

	Figures~\ref{fig:predicate_sentiment_distribution} and~\ref{fig:predicate_sentiment_buckets} 
	illustrate the sentiment score distribution and the corresponding class frequencies.
	
	As shown, most predicates (approximately 87.6\%) are neutral, 
	reflecting the factual style of WSJ writing that minimizes overt emotional tone. 
	The remaining classes are balanced, each around 3\% of the total. 
	This neutrality aligns with the corpus’s journalistic nature and ensures that extracted structures 
	serve as unbiased inputs for downstream \gls*{srl} and \gls*{orl} modeling.

	\section{Conclusion}
	
	This technical report introduces a transparent, fully reproducible pipeline for extracting \numInstances high-quality \gls*{srl} instances from OntoNotes 5.0 (WSJ), explicitly aligned with \gls*{orl} roles. By resolving syntactic tree pointers, handling discontinuous arguments, correcting annotation errors, and rigorously cleaning trace markers, we produce a robust, ready-to-use dataset for low-resource opinion mining.
	
	The complete extraction code, preprocessing scripts, and final dataset are publicly released, enabling direct application in transfer learning, multi-task learning, and active learning frameworks. This work establishes a reliable bridge between semantic and opinion role labeling, offering a reusable resource for advancing \gls*{orl} in data-scarce domains.
	
	Future directions include cross-domain evaluation (e.g., social media), integration with generative ORL models (e.g., GenOM), and extension to multilingual settings.

	\bibliographystyle{unsrt}  
	
	\bibliography{anthology,custom}

\end{document}